# Generalized Correntropy for Robust Adaptive Filtering

Badong Chen[1], Lei Xing[1], Haiquan Zhao[2], Nanning Zheng[1], José C. Príncipe[1,3]

*1. Institute of Artificial Intelligence and Robotics, Xi'an Jiaotong University, Xi'an, 710049, China*
*2. School of Electrical Engineering, Southwest Jiaotong University, Chengdu, China*
*3. Department of Electrical and Computer Engineering, University of Florida, Gainesville, FL 32611 USA*

***Abstract*—As a robust nonlinear similarity measure in kernel space, correntropy has received increasing attention in domains of machine learning and signal processing. In particular, the maximum correntropy criterion (MCC) has recently been successfully applied in robust regression and filtering. The default kernel function in correntropy is the Gaussian kernel, which is, of course, not always the best choice. In this work, we propose a generalized correntropy that adopts the generalized Gaussian density (GGD) function as the kernel (not necessarily a Mercer kernel), and present some important properties. We further propose the generalized maximum correntropy criterion (GMCC), and apply it to adaptive filtering. An adaptive algorithm, called the *GMCC algorithm*, is derived, and the mean square convergence performance is studied. We show that the proposed algorithm is very stable and can achieve zero probability of divergence (POD). Simulation results confirm the theoretical expectations and demonstrate the desirable performance of the new algorithm.**

***Key Words:*** *Correntropy; generalized correntropy; adaptive filtering, GMCC algorithm.*

## I. INTRODUCTION

Selecting a proper cost function (usually a statistical measure of the error signals) is a key issue in adaptive filtering theory and applications [1-3]. The mean square error (MSE) is widely used as a cost function since it has attractive features, such as smoothness, convexity, mathematical tractability, low computational burden and optimality under Gaussian assumption. The well-known least mean square (LMS) algorithm and its variants, such as normalized LMS (NLMS) and variable step-size LMS (VSSLMS), were developed under this criterion [1, 2]. The MSE is desirable if the signals are Gaussian distributed. In non-Gaussian situations, however, its performance may degrade considerably and in these cases, a non-quadratic cost will be, in general, better than MSE [3].

Generally speaking, there are two types of non-Gaussian distributions: light-tailed (e.g. uniform, binary, etc.) and heavy-tailed (e.g. Laplace, $\alpha$ -stable, etc.) distributions. When the desired signals are disturbed by light-tailed non-Gaussian noises, a higher-order statistical (HOS) measure of the error is usually more desirable. A typical example is the least mean fourth (LMF) family algorithms, which use the mean even power of the error as the cost function [4]. Compared with the LMS algorithm, the LMF may achieve a faster convergence speed and a lower steady-state mean square deviation (MSD) especially in light-tailed noises. One drawback of the LMF algorithm however is that the stability is not guaranteed, which depends on the input and noise powers, and the initial values of the weights. A more general class of algorithms are the least mean $p$ -power (LMP) family algorithms, which adopt the $p$ -order absolute moment of the error as the adaptation cost [5].

When the desired signals are disturbed by heavy-tailed impulsive noises (which may cause large outliers), a lower-order statistical (LOS) measure of the error is usually more robust (i.e. less sensitive to impulsive interferences). For example, the sign algorithm (SA), which employs the mean absolute value of the error as the cost function, is rather robust to the presence of large noises [6-8]. The convergence speed and steady-state performance of the SA algorithm is however not so good in general. In the literature many other robust cost functions have been proposed to develop robust adaptive filtering algorithms. Typical examples include mixed-norm [9, 10], M-estimation cost [11,12], and error entropy [13-18]. Particularly in recent years, the maximum correntropy criterion (MCC) has been successfully used in robust adaptive filtering, wherein the filter weights are adapted such that the correntropy between the desired signal and filter output is maximized [19-24]. The correntropy is a nonlinear and local similarity measure directly related to the probability of how similar two random variables are in a neighborhood of the joint space controlled by the kernel bandwidth, which also has its root in Renyi's entropy (hence the name

"correntropy") [13, 19]. Since correntropy is insensitive to outliers especially with a small kernel bandwidth, it is naturally a robust adaptation cost in presence of heavy-tailed impulsive noises.

The kernel function in correntropy is usually a Gaussian kernel, which is desirable due to its smoothness and *strict positive-definiteness*. With a Gaussian kernel, the correntropy induces a nonlinear metric called the *correntropy induced metric* (CIM) which behaves like an $L_2$ norm when data are relatively small compared with the kernel bandwidth, an $L_1$ norm as data get larger, and an $L_0$ norm when data are far away from the origin [19, 25]. However, Gaussian kernel is, of course, not always the best choice. In the present work, we propose to use the *generalized Gaussian density* (GGD) [26, 27] function (which is not necessarily a Mercer kernel) as a kernel function in correntropy, and the new correntropy is called the *generalized correntropy*. Some important properties of the generalized correntropy are presented. In particular, we show that the order-$\alpha$ *generalized correntropy induced metric* (GCIM) or *generalized correntropic loss* (GC-loss) function behaves like different norms (from $L_\alpha$ to $L_0$) of the data in different regions.

Similar to the original correntropy with Gaussian kernel, the generalized correntropy can also be used as an optimization cost in estimation-related problems. It can be proven that in essence the *generalized maximum correntropy criterion* (GMCC) based estimation is a smoothed maximum *a posteriori* probability (MAP) estimation, including the MAP and the *least mean* $p$ *-power* (LMP) estimation as the extreme cases. In this work, we focus mainly on applying the GMCC criterion to adaptive filtering. We show that the optimal solution of GMCC filtering is in form similar to the well-known Wiener solution, except that the autocorrelation matrix and cross-correlation vector are weighted by an error nonlinearity. If the signals involved are zero-mean Gaussian, the optimal solution will equal to the Wiener solution. Under the GMCC criterion, a stochastic gradient based adaptive filtering algorithm, called the *GMCC algorithm*, is developed. The mean-square convergence performance of the GMCC algorithm is analyzed. In particular, we present a simple example to show that the GMCC will have a zero probability of divergence (POD). A theoretical value of the steady-state *excess mean square error* (EMSE) of the GMCC algorithm is also derived. Simulation results confirm the theoretical expectations and the desirable performance of the GMCC.

The rest of the paper is organized as follows. In section II, we define the generalized correntropy, and present some important properties. In section III, we propose the generalized maximum correntropy criterion (GMCC). In section IV, we apply the GMCC criterion to adaptive filtering and develop the GMCC algorithm. In section V, we analyze the mean square convergence of the GMCC algorithm. In section VI, we present Monte Carlo simulation results to verify the theoretical results and demonstrate the performance of the new algorithm. Finally in section VII, we give the conclusion.

## II. Generalized Correntropy

### *A. Definition*

Given two random variables $X$ and $Y$, the *correntropy* is defined by [19, 28]

$$\mathrm{V}(X,Y)=\mathbf{E}[\kappa(X,Y)]=\int\kappa(x,y)dF_{XY}(x,y) \tag{1}$$

where $\mathbf{E}$ denotes the expectation operator, $\kappa(\cdot,\cdot)$ is a *shift-invariant Mercer kernel*, and $F_{XY}(x,y)$ denotes the joint distribution function of $(X,Y)$. Without mentioned otherwise, the kernel function of correntropy is the Gaussian kernel:

$$\kappa(x,y)=\mathrm{G}_\sigma(e)=\frac{1}{\sqrt{2\pi}\sigma}\exp\left(-\frac{e^2}{2\sigma^2}\right)=\frac{1}{\sqrt{2\pi}\sigma}\exp\left(-\lambda e^2\right) \tag{2}$$

where $e=x-y$, $\sigma>0$ is the kernel bandwidth, and $\lambda=-1/2\sigma^2$ is the kernel parameter. The correntropy $\mathrm{V}(X,Y)$ can also be expressed as

$$\mathrm{V}(X,Y)=\mathbf{E}\left[\varphi(X)^T\varphi(Y)\right]=trace\left(\mathbf{R}_{\varphi(X)\varphi(Y)}\right) \tag{3}$$

where $\mathbf{R}_{\varphi(X)\varphi(Y)} = \mathbf{E}\left[\varphi(X)\varphi(Y)^T\right]$, and $\varphi(.)$ denotes a nonlinear mapping induced by $\kappa$, which transforms its argument into a high-dimensional (infinite for Gaussian kernels) Hilbert space $\mathcal{F}_\kappa$, satisfying $\varphi(X)^T\varphi(Y) = \kappa(X,Y)$ [19]. The correntropy is therefore essentially a second-order statistic of the mapped feature space data.

There is a well-known generalization of Gaussian density function called the *generalized Gaussian density* (GGD) function, which with zero-mean is given by [26, 27]

$$\mathrm{G}_{\alpha,\beta}(e) = \frac{\alpha}{2\beta\Gamma(1/\alpha)}\exp\left(-\left|\frac{e}{\beta}\right|^{\alpha}\right) = \gamma_{\alpha,\beta}\exp\left(-\lambda|e|^{\alpha}\right) \tag{4}$$

where $\Gamma(.)$ is the gamma function, $\alpha > 0$ is the shape parameter, $\beta > 0$ is the scale (bandwidth) parameter, $\lambda = 1/\beta^{\alpha}$ is the kernel parameter, and $\gamma_{\alpha,\beta} = \alpha/\left(2\beta\Gamma(1/\alpha)\right)$ is the normalization constant. This parametric family of symmetric distributions include the Gaussian ( $\alpha = 2$ ) and Laplace ( $\alpha = 1$ ) distributions as the special cases. Further, as $\alpha \to \infty$, the GGD density converges point-wise to a uniform density on $(-\beta, \beta)$.

In this work, we use the GGD density function as the kernel function of correntropy, and define

$$\mathrm{V}_{\alpha,\beta}(X,Y) = \mathbf{E}[\mathrm{G}_{\alpha,\beta}(E_{X-Y})] = \mathbf{E}[\mathrm{G}_{\alpha,\beta}(X-Y)] \tag{5}$$

where $E_{X-Y} = X - Y$ .To make a distinction between (5) and the correntropy with Gaussian kernel, we call it the generalized correntropy. Clearly, the correntropy with Gaussian kernel corresponds to the generalized correntropy with $\alpha = 2$.

*Remark 1*: It is worth noting that in the generalized correntropy, the kernel function does not necessarily satisfy the Mercer's condition. Actually, the kernel function $\kappa(x,y) = \mathrm{G}_{\alpha,\beta}(x-y)$ is positive definite if and only if $0 < \alpha \le 2$ (see [29] page 434).

In practice, the joint distribution of $X$ and $Y$ is usually unknown, and only a finite number of samples $\{(x_i, y_i)\}_{i=1}^{N}$ are available. In this case, the sample estimator of the generalized correntropy is

$$\hat{\mathrm{V}}_{\alpha,\beta}(X,Y) = \frac{1}{N}\sum_{i=1}^{N}\mathrm{G}_{\alpha,\beta}(x_i - y_i) \tag{6}$$

*B.Properties*

Below we present several basic properties of the generalized correntropy. Some of them are simple extensions of the properties presented in [19], and will not be proved here.

*Property 1*: $\mathrm{V}_{\alpha,\beta}(X,Y)$ is symmetric, that is $\mathrm{V}_{\alpha,\beta}(X,Y) = \mathrm{V}_{\alpha,\beta}(Y,X)$.

*Property 2*: $\mathrm{V}_{\alpha,\beta}(X,Y)$ is positive and bounded: $0 < \mathrm{V}_{\alpha,\beta}(X,Y) \le \mathrm{G}_{\alpha,\beta}(0) = \gamma_{\alpha,\beta}$, and it reaches its maximum if and only if $X = Y$.

*Property 3*: The generalized correntropy involves higher-order absolute moments of the error variable $E_{X-Y} = X - Y$: $\mathrm{V}_{\alpha,\beta}(X,Y) = \gamma_{\alpha,\beta}\sum_{n=0}^{\infty}\frac{(-\lambda)^n}{n!}\mathbf{E}\left[|X-Y|^{\alpha n}\right]$.

*Remark 2*: When the kernel parameter $\lambda$ is small enough, we have $\mathrm{V}_{\alpha,\beta}(X,Y) \approx \gamma_{\alpha,\beta}\left(1 - \lambda\mathbf{E}\left[|X-Y|^{\alpha}\right]\right)$. In this case, the generalized correntropy is, approximately, an affine linear function of the $\alpha$ -order absolute moment of the error $E_{X-Y}$.

*Property 4*: Assume that the samples $\{(x_i, y_i)\}_{i=1}^{N}$ are drawn from the joint PDF $p_{XY}(x, y)$. Let $\hat{p}_E(e)$ be the Parzen estimate of the error PDF from samples $\{e_i = x_i - y_i\}_{i=1}^{N}$, with the GGD density function $\mathrm{G}_{\alpha,\beta}$ as the Parzen window kernel. Then $\hat{\mathrm{V}}_{\alpha,\beta}(X,Y)$ is the value of $\hat{p}_E(e)$ evaluated at the point $e = 0$, that is

$$\hat{\mathrm{V}}_{\alpha,\beta}(X,Y) = \hat{p}_E(0) \tag{7}$$

where $\hat{p}_E(e) = \frac{1}{N}\sum_{i=1}^{N} \mathrm{G}_{\alpha,\beta}(e - e_i)$ .

*Property 5*: For the case $0 < \alpha \le 2$, the generalized correntropy is a second-order statistic of the mapped feature space data.

*Proof*: When $0 < \alpha \le 2$, the kernel function $\kappa(x, y) = \mathrm{G}_{\alpha,\beta}(x - y)$ is a Mercer kernel, and hence we have $\mathrm{V}_{\alpha,\beta}(X,Y) = \mathbf{E}\left[\varphi_{\alpha,\beta}(X)^T \varphi_{\alpha,\beta}(Y)\right]$, where $\varphi_{\alpha,\beta}(.)$ is a nonlinear mapping induced by $\mathrm{G}_{\alpha,\beta}$ .

In data analysis such as regression and classification, a measure called the *correntropic loss* (C-loss) is usually used instead of using the correntropy [30, 31]. A generalized C-loss (GC-loss) function between $X$ and $Y$ can be defined as

$$J_{GC-loss}(X,Y) = \mathrm{G}_{\alpha,\beta}(0) - \mathrm{V}_{\alpha,\beta}(X,Y) \tag{8}$$

The GC-loss satisfies $J_{GC-loss}(X,Y) \ge 0$, and when $0 < \alpha \le 2$, it can be expressed as

$$J_{GC-loss}(X,Y) = \frac{1}{2}\mathbf{E}\left[\left\|\varphi_{\alpha,\beta}(X) - \varphi_{\alpha,\beta}(Y)\right\|^2\right] \tag{9}$$

which is a mean-square loss in the feature space $\mathcal{F}_\kappa$ induced by Mercer kernel $\kappa(x, y) = \mathrm{G}_{\alpha,\beta}(x - y)$. Clearly, minimizing the GC-loss will be equivalent to maximizing the generalized correntropy.

Let $\{(x_i, y_i)\}_{i=1}^N$ be the samples drawn from $p_{XY}$ . Then an estimator of the GC-loss is

$$\begin{aligned}\hat{J}_{GC-loss}(X,Y) &= \mathrm{G}_{\alpha,\beta}(0) - \hat{\mathrm{V}}_{\alpha,\beta}(X,Y) \\ &= \gamma_{\alpha,\beta} - \frac{1}{N}\sum_{i=1}^{N} \mathrm{G}_{\alpha,\beta}(x_i - y_i) \\ &= \gamma_{\alpha,\beta} - \frac{1}{N}\sum_{i=1}^{N} \mathrm{G}_{\alpha,\beta}(e_i)\end{aligned} \tag{10}$$

*Property 6*: Let $\vec{X}=\left[x_1,\cdots,x_N\right]^T$, $\vec{Y}=\left[y_1,\cdots,y_N\right]^T$. Then the function $GCIM\left(\vec{X},\vec{Y}\right)=\sqrt{\hat{J}_{GC-loss}(X,Y)}$ , called the *generalized correntropy induced metric* (GCIM), defines a metric in the $N$ -dimensional sample vector space when $0<\alpha\le 2$.

*Proof*: See Appendix A.

For the case in which $\vec{X}=\left[x_1,x_2\right]^T$, $\vec{Y}=\left[0,0\right]^T$, $\alpha=4$, and $\lambda=1$, the surface of the $GCIM\left(\vec{X},\vec{Y}\right)$ with respect to $x_1$ and $x_2$ is shown in Fig. 1. As one can see, the GCIM behaves like different norms (from $L_\alpha$ to $L_0$) of the data in different regions. This observation, which is similar to that obtained with Gaussian kernel [19], is confirmed by Properties 7 and 8.

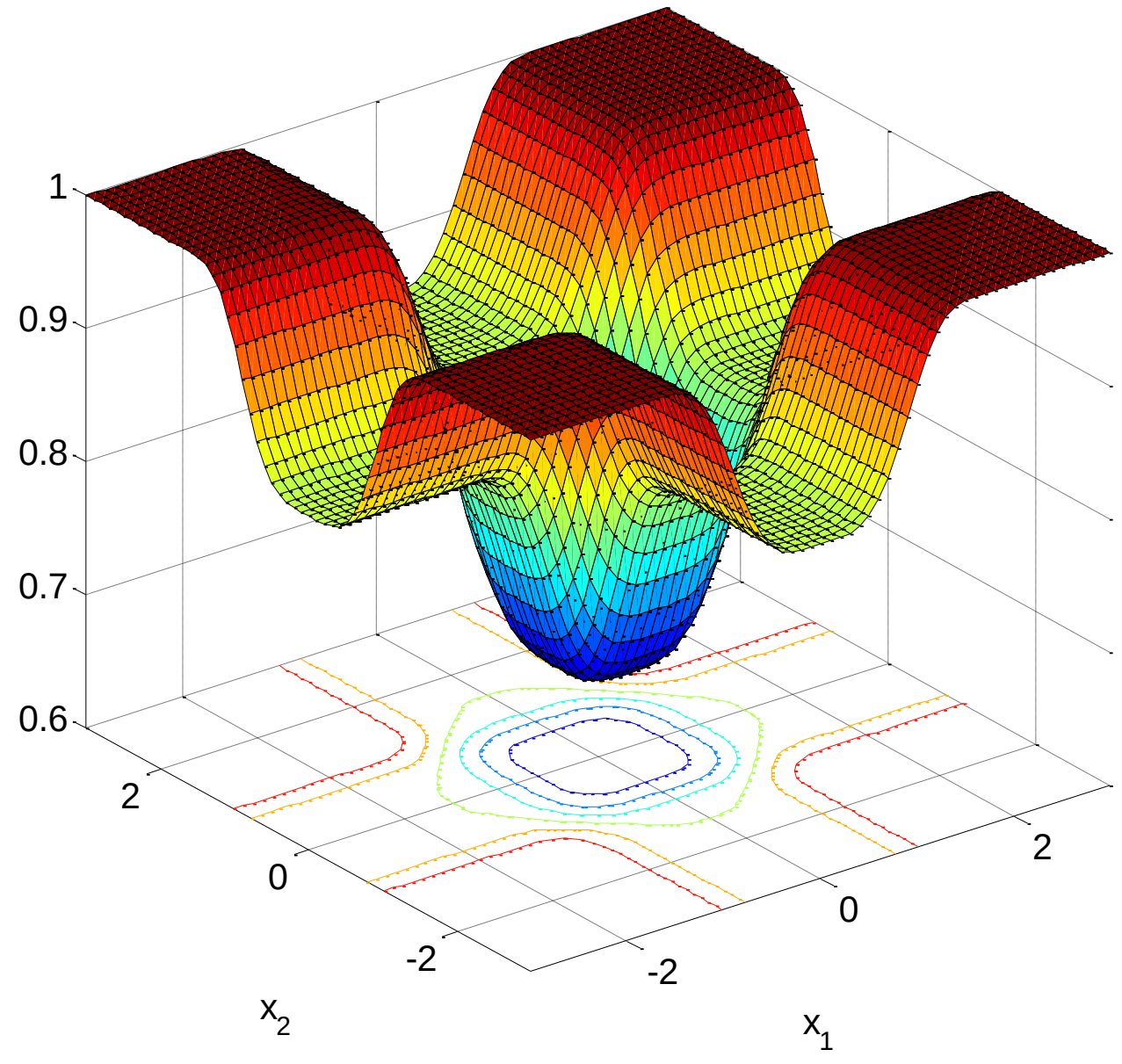

Fig.1 Surface of the GCIM in 3D space ( $\alpha=4$, $\lambda=1$ ).

*Property 7*: As $\lambda\to 0+$ (or $x_i\to 0, i=1,\cdots,N$ ), the function $L_{\alpha,\beta}\left(\vec{X}\right)=\left(\frac{N}{\lambda\gamma_{\alpha,\beta}}\hat{J}_{GC-loss}(X,0)\right)^{1/\alpha}$ will approach the $l_\alpha$ -norm of $\vec{X}$ , that is

$$L_{\alpha,\beta}\left(\vec{X}\right) \approx \left\|\vec{X}\right\|_{\alpha} = \left(\sum_{i=1}^{N} |x_i|^{\alpha}\right)^{1/\alpha}, \text{ as } \lambda \to 0 \tag{11}$$

*Proof*: See Appendix B.

*Property 8*: Assume that $|x_i| > \delta$, $\forall i : x_i \neq 0$, where $\delta$ is a small positive number. As $\lambda \to \infty$ (or $\beta \to 0+$), minimizing the function $L_{\alpha,\beta}\left(\vec{X}\right)$ will be, approximately, equivalent to minimizing the $l_0$-norm of $\vec{X}$, that is

$$\min_{\vec{X}\in\Omega} L_{\alpha,\beta}\left(\vec{X}\right) \sim \min_{\vec{X}\in\Omega} \left\|\vec{X}\right\|_0, \text{ as } \lambda \to \infty \tag{12}$$

where $\Omega$ denotes a feasible set of $\vec{X}$.

*Proof*: Similar to the one presented in [32]. See Appendix C for the detailed derivation.

Below we present some optimization properties of the GC-loss. Similar results for the C-loss can be found in [31].

*Property 9:* Let $\vec{e} = \left[e_1, \cdots, e_N\right]^T$. Then the following statements hold:

1) if $0 < \alpha \leq 1$, then the GC-loss $\hat{J}_{GC-loss}$ is concave at any $\vec{e}$ with $e_i \neq 0$ ($i = 1, \cdots, N$);

2) if $\alpha > 1$, then the GC-loss $\hat{J}_{GC-loss}$ is convex at any $\vec{e}$ with $0 < |e_i| \leq \left[(\alpha-1)/\alpha\lambda\right]^{1/\alpha}$ ($i = 1, \cdots, N$);

3) if $\lambda \to 0+$, then for any $\vec{e}$ with $e_i \neq 0$ ($i = 1, \cdots, N$), the GC-loss $\hat{J}_{GC-loss}$ is concave at $\vec{e}$ for $0 < \alpha \leq 1$, and convex at $\vec{e}$ for $\alpha > 1$.

*Proof*: See Appendix D.

*Property 10:* For $\alpha > 1$, the GC-loss $\hat{J}_{GC-loss}$ is a differentiable invex function of $\vec{e} = \left[e_1, \cdots, e_N\right]^T$ with $e_i \leq M$ ($i = 1, \cdots, N$), where $M$ is an arbitrary positive number.

*Proof*: See Appendix E.

## III. Generalized Maximum Correntropy Criterion

Similar to the correntropy, the generalized correntropy can also be used as an optimization cost in estimation-related problems. Given two random variables: $X \in \mathbb{R}$, an unknown real-valued parameter to be estimated, $Y \in \mathbb{R}^m$, a vector of observations (measurements), an estimator of $X$ can be defined as a function of $Y$ : $\hat{X} = g(Y)$, where $g$ is solved by optimizing a certain cost function. Under the *generalized maximum correntropy criterion* (GMCC), the estimator $g$ will be solved by maximizing the generalized correntropy between $X$ and $\hat{X}$ , that is

$$g_{GMCC} = \arg\max_{g \in \mathbf{G}} \mathrm{V}_{\alpha,\beta}(X, \hat{X}) = \arg\max_{g \in \mathbf{G}} \mathbf{E}[\mathrm{G}_{\alpha,\beta}(X - g(Y))] \tag{13}$$

where $\mathbf{G}$ stands for the collection of all measurable functions of $Y$ . If $\forall y \in \mathbb{R}^m$, $X$ has a conditional PDF $p_{X|Y}(x|y)$, then the estimation error $E_{X-\hat{X}} = X - \hat{X}$ has PDF

$$p_E(e) = \int_{\mathbb{R}^m} p_{X|Y}(e + g(y) \mid y) dF_Y(y) \tag{14}$$

where $F_Y(y)$ is the distribution function of $Y$ . It follows that

$$g_{GMCC} = \arg\max_{g \in \mathbf{G}} \int_{\mathbb{R}} \mathrm{G}_{\alpha,\beta}(e) \int_{\mathbb{R}^m} p_{X|Y}(e + g(y) \mid y) dF_Y(y) de \tag{15}$$

In [28], it has been proven that the MCC estimate is a smoothed maximum *a posteriori* probability (MAP) estimate, which equals the mode (at which the PDF attains its maximum value) of a smoothed *a posteriori* distribution. A similar result holds for the GMCC estimation.

*Theorem 1:* The GMCC estimator can be expressed as

$$g_{GMCC}(y) = \arg\max_{x \in \mathbb{R}} \rho_{\alpha,\beta}(x \mid y), \ \forall y \in \mathbb{R}^m \tag{16}$$

where $\rho_{\alpha,\beta}(x \mid y) = G_{\alpha,\beta}(x) * p_{X|Y}(x \mid y)$ ("$*$"denotes the convolution operator with respect to $x$ ).

*Proof:* See Appendix F.

*Remark 3:* Considering the conditional PDF $p_{X|Y}(x|y)$ as an *a posteriori* PDF given the observation $y$, the function $\rho_{\alpha,\beta}(x\,|\,y)$ will be a smoothed (by convolution) *a posteriori* PDF. Therefore, the GMCC estimate is a smoothed MAP estimate.

*Corollary 1:* When $\beta \to 0+$ (or $\lambda \to \infty$), the GMCC estimation becomes the MAP estimation.

*Proof:* As $\beta \to 0+$, the GGD function $\mathrm{G}_{\alpha,\beta}(.)$ will approach the Dirac delta function, and the smoothed conditional PDF $\rho_{\alpha,\beta}(x\,|\,y)$ will reduce to the original conditional PDF $p_{X|Y}(x|y)$. In this case, the GMCC estimation will be equivalent to the MAP estimation.

*Theorem 2:* The GMCC estimator can also be expressed as

$$g_{GMCC} = \arg\max_{g \in \mathbf{G}} p_E^{\alpha,\beta}(0) \tag{17}$$

where $p_E^{\alpha,\beta}(x) = p_E(x) * \mathrm{G}_{\alpha,\beta}(x)$ is the smoothed error PDF.

*Proof:* It is easy to derive

$$\begin{aligned} \mathbf{E}\left[\mathrm{G}_{\alpha,\beta}(E_{X-\hat{X}})\right] &= \int_{\mathbb{R}} \mathrm{G}_{\alpha,\beta}(e) p_E(e) de \\ &= \left(\int_{\mathbb{R}} \mathrm{G}_{\alpha,\beta}(x-e) p_E(e) de\right)_{x=0} \\ &= \left(p_E(x) * \mathrm{G}_{\alpha,\beta}(x)\right)_{x=0} \\ &= p_E^{\alpha,\beta}(0) \end{aligned} \tag{18}$$

And hence

$$g_{GMCC} = \arg\max_{g \in \mathbf{G}} \mathbf{E}\left[\mathrm{G}_{\alpha,\beta}(E_{X-\hat{X}})\right] = \arg\max_{g \in \mathbf{G}} p_E^{\alpha,\beta}(0) \tag{19}$$

*Remark 4:* Theorem 2 implies that the GMCC estimator will make the error distribution concentrated around zero with a high peak.

*Theorem 3:* When $\beta \to \infty$ (or $\lambda \to 0+$), the GMCC estimation will be equivalent to the *least mean $p$-power* (LMP) estimation with $p = \alpha$.

*Proof:* Under the LMP criterion, the optimal estimator is solved by minimizing the mean $p$ -power of the error:

$$g_{LMP} = \arg\min_{g \in \mathbf{G}} \mathbf{E}\left[\left|E_{X-\hat{X}}\right|^p\right] = \arg\min_{g \in \mathbf{G}} \mathbf{E}\left[\left|X - \hat{X}\right|^p\right] \tag{20}$$

On the other hand, as $\lambda \to 0+$, we have $\mathrm{V}_{\alpha,\beta}(X,\hat{X}) \approx \gamma_{\alpha,\beta}\left(1-\lambda\mathbf{E}\left[\left|X-\hat{X}\right|^\alpha\right]\right)$. It follows that

$$\max \mathrm{V}_{\alpha,\beta}(X,\hat{X}) \sim \min \mathbf{E}\left[\left|X-\hat{X}\right|^\alpha\right] \tag{21}$$

That is, as $\lambda \to 0+$, the GMCC estimation will be, approximately, equivalent to the LMP estimation with $p = \alpha$.

## IV. Adaptive Filtering under GMCC Criterion

### *A. Cost function*

In the context of linear adaptive filtering, under the GMCC criterion the optimal weight vector of the filter can be solved by maximizing

$$J_{GMCC} = \mathbf{E}\left[\mathrm{G}_{\alpha,\beta}(e(i))\right] = \gamma_{\alpha,\beta}\mathbf{E}\left[\exp\left(-\lambda\left|e(i)\right|^\alpha\right)\right] \tag{22}$$

where the error

$$e(i) = d(i) - y(i) = d(i) - W^T X(i) \tag{23}$$

with $d(i) \in \mathbb{R}$ being the desired value at time $i$, $y(i) = W^T X(i)$ the output of the filter, $W = \left[w_1, w_2, \cdots, w_m\right]^T \in \mathbb{R}^m$ the weight vector, and $X(i) \in \mathbb{R}^m$ the input vector, generally given by

$$X(i) = \left[x(i), x(i-1), \cdots, x(i-m+1)\right]^T \tag{24}$$

where $x(i)$ is the input signal.

*B. Optimal solution*

On the optimal weight vector of GMCC adaptive filtering, we summarize the main result as the following theorem.

*Theorem 4:* The optimal weight vector that maximizes $J_{GMCC}$ can be expressed as

$$W_{opt} = \left[\mathbf{R}_{XX}^{h}\right]^{-1}\mathbf{P}_{dX}^{h} \tag{25}$$

where $\mathbf{R}_{XX}^{h} = \mathbf{E}\left[h\left(e(i)\right)X(i)X(i)^{T}\right]$ is a weighted autocorrelation matrix of the input signal, in which the weighting is a function of the error $h(e(i)) = \exp\left(-\lambda\left|e(i)\right|^{\alpha}\right)\left|e(i)\right|^{\alpha-2}$, and $\mathbf{P}_{dX}^{h} = \mathbf{E}\left[h\left(e(i)\right)d(i)X(i)\right]$ is a weighted cross-correlation vector between the desired and the input vector.

*Proof:* Let $\frac{\partial}{\partial W}J_{GMCC} = 0$, we have

$$\begin{aligned}
&\mathbf{E}\left[\exp\left(-\lambda\left|e(i)\right|^{\alpha}\right)\left|e(i)\right|^{\alpha-1} sign\left(e(i)\right)X(i)\right] = 0\\
&\Rightarrow \mathbf{E}\left[h\left(e(i)\right)\left(d(i) - W^{T}X(i)\right)X(i)\right] = 0\\
&\Rightarrow \mathbf{E}\left[h\left(e(i)\right)X(i)X(i)^{T}\right]W = \mathbf{E}\left[h\left(e(i)\right)d(i)X(i)\right]\\
&\Rightarrow W = \left[\mathbf{R}_{XX}^{h}\right]^{-1}\mathbf{P}_{dX}^{h}
\end{aligned} \tag{26}$$

*Remark 5:* For the case $\alpha = 2$, we have $h(e(i)) = \exp\left(-\lambda e^{2}(i)\right)$. In this case, as $\lambda \to 0+$, we have $h(e(i)) \to 1$, and $W_{opt} \to \mathbf{R}_{XX}^{-1}\mathbf{P}_{dX}$ with $\mathbf{R}_{XX} = \mathbf{E}\left[X(i)X(i)^{T}\right]$, $\mathbf{P}_{dX} = \mathbf{E}\left[d(i)X(i)\right]$, which corresponds to the well-known *Wiener solution*.

*Theorem 5:* If $x(i)$ and $d(i)$ are both zero-mean Gaussian processes, then the optimal solution under GMCC criterion is equal to the Wiener solution.

*Proof:* First, we derive

$$\begin{aligned}\mathbf{E}\left[\mathrm{G}_{\alpha,\beta}\left(e(i)\right)\right] &= \int_{\mathbb{R}} \mathrm{G}_{\alpha,\beta}(e)\, p_{e(i)}(e)de \\ &= \int_{\mathbb{R}} p_{e(i)}(e)\left(\int_0^{\gamma_{\alpha,\beta}} \mathbb{I}\left(\xi \le \mathrm{G}_{\alpha,\beta}(e)\right)d\xi\right)de \\ &= \int_0^{\gamma_{\alpha,\beta}}\left(\int_{-\infty}^{\infty} p_{e(i)}(e)\mathbb{I}\left(\xi \le \mathrm{G}_{\alpha,\beta}(e)\right)de\right)d\xi \\ &= \int_0^{\gamma_{\alpha,\beta}}\left(\int_{\left\{e:\mathrm{G}_{\alpha,\beta}(e)\ge\xi\right\}} p_{e(i)}(e)de\right)d\xi \\ &= \int_0^{\gamma_{\alpha,\beta}}\left(\int_{-\varepsilon}^{\varepsilon} p_{e(i)}(e)de\right)d\xi\end{aligned} \tag{27}$$

where $\mathbb{I}(.)$ denotes an indicator function, and $\varepsilon$ is a certain positive number satisfying $\mathrm{G}_{\alpha,\beta}(\varepsilon) = \mathrm{G}_{\alpha,\beta}(-\varepsilon) = \xi$ .Since $x(i)$ and $d(i)$ are both zero-mean Gaussian processes, then the error $e(i)$ is also a zero-mean Gaussian process, with PDF

$$p_{e(i)}(e) = \frac{1}{\sqrt{2\pi}\sigma_{e(i)}}\exp\left(-\frac{e^2}{2\sigma_{e(i)}^2}\right) \tag{28}$$

where $\sigma_{e(i)}^2 = \mathbf{E}\left[e^2(i)\right]$ is the error variance. Hence

$$\int_{-\varepsilon}^{\varepsilon} p_{e(i)}(e)de = \operatorname{erf}\left(\frac{\varepsilon}{\sqrt{2}\sigma_{e(i)}}\right) \tag{29}$$

with $\operatorname{erf}(x) = \frac{1}{\sqrt{\pi}}\int_{-x}^{x}\exp(-t^2)dt$ being the *error function*, which is a monotonically increasing function of $x$ . Therefore, $\int_{-\varepsilon}^{\varepsilon} p_{e(i)}(e)de$ is a monotonically decreasing function of $\sigma_{e(i)}^2$ . It follows that

$$\max \mathbf{E}\left[\mathrm{G}_{\alpha,\beta}\left(e(i)\right)\right] \Leftrightarrow \max\int_{-\varepsilon}^{\varepsilon} p_{e(i)}(e)de \Leftrightarrow \min\sigma_{e(i)}^2 \tag{30}$$

That is, the maximization of the generalized correntropy $\mathbf{E}\left[\mathrm{G}_{\alpha,\beta}\left(e(i)\right)\right]$ will be equivalent to the minimization of the mean square error $\sigma_{e(i)}^2 = \mathbf{E}\left[e^2(i)\right]$. Clearly, in this case, the optimal solution under GMCC criterion will be equal to the Wiener solution $\mathbf{R}_{XX}^{-1}\mathbf{P}_{dX}$ .

*C.Adaptive algorithm*

Based on the cost function of (22), a stochastic gradient based adaptive algorithm, called in this work the *GMCC algorithm*, can be simply derived as

$$\begin{aligned} W(i+1) &= W(i) + \mu \frac{\partial}{\partial W(i)} \exp\left(-\lambda |e(i)|^{\alpha}\right) \\ &= W(i) - \mu\lambda\alpha \exp\left(-\lambda |e(i)|^{\alpha}\right) |e(i)|^{\alpha-1} sign(e(i)) \frac{\partial e(i)}{\partial W(i)} \\ &= W(i) + \eta \exp\left(-\lambda |e(i)|^{\alpha}\right) |e(i)|^{\alpha-1} sign(e(i)) X(i) \end{aligned} \quad (31)$$

where $\eta = \mu\lambda\alpha$ is the step-size parameter.

We have the following observations:

1) When $\alpha = 2$, the GMCC algorithm becomes

$$W(i+1) = W(i) + \eta \exp\left(-\lambda e^2(i)\right) e(i) X(i) \quad (32)$$

which is the original MCC algorithm [20].

2) The weight update equation of (31) can be rewritten as

$$W(i+1) = W(i) + \eta(i) |e(i)|^{\alpha-1} sign(e(i)) X(i) \quad (33)$$

where $\eta(i) = \eta \exp\left(-\lambda |e(i)|^{\alpha}\right)$. Therefore, the GMCC algorithm can be viewed as an LMP algorithm with $p = \alpha$ and a variable step-size $\eta(i)$. The LMP algorithm is derived under the LMP (*least mean* $p$ *-power*) criterion, which includes SA ( $p = 1$), LMS ( $p = 2$ ), and LMF ( $p = 4$ ) as special cases (see [5] for the details).

3) When $\lambda \to 0+$, we have $\eta(i) \to \eta$. In this case, the GMCC algorithm reduces to the traditional LMP algorithm with $p = \alpha$ :

$$W(i)=W(i-1)+\eta\left|e(i)\right|^{\alpha-1}sign(e(i))X(i) \tag{34}$$

In particular, when $\alpha=2$, (34) becomes the well-known LMS algorithm:

$$W(i)=W(i-1)+\eta e(i)X(i) \tag{35}$$

4) When $\left|e(i)\right|\to\infty$, we have $\eta(i)\to 0$. Thus, a large error will have little influence on the filter weights. This implies that the GMCC algorithm will be robust to large outliers (or impulsive noises), which often cause large errors.

*Remark 6*: One can further derive various variants of the GMCC algorithm, such as the variable step-size GMCC and normalized GMCC, where the step-size is changed across iterations or divided by the squared norm of the input vector.

The computational complexity of the GMCC algorithm is almost the same as the LMP algorithm, and the only extra computational effort needed is to calculate the term $\exp\left(-\lambda\left|e(i)\right|^{\alpha}\right)$, which is obviously not expensive.

## V. Mean-Square Convergence Analysis

In this section, we analyze the mean-square convergence performance of the proposed GMCC algorithm. The algorithm (31) can be written in a general form:

$$W(i)=W(i-1)+\eta f\left(e(i)\right)X(i) \tag{36}$$

where $f\left(e(i)\right)$ is a nonlinear function of $e(i)$,

$$f\left(e(i)\right)=\exp\left(-\lambda\left|e(i)\right|^{\alpha}\right)\left|e(i)\right|^{\alpha-1}sign(e(i)) \tag{37}$$

Assume that the desired signal $d(i)$ can be expressed as

$$d(i)=W_0^{T}X(i)+v(i) \tag{38}$$

where $W_0$ denotes an unknown weight vector that needs to be estimated, and $v(i)$ stands for the disturbance noise, with variance $\sigma_v^2$. Then, we have

$$e(i)=\tilde{W}(i-1)^T X(i)+v(i)=e_a(i)+v(i) \tag{39}$$

where $\tilde{W}(i-1)=W_0-W(i-1)$ is the *weigh error vector* at iteration $i-1$, and $e_a(i)=\tilde{W}(i-1)^T X(i)$ is referred to as the *a priori* error. After some simple algebra, one can obtain [33]:

$$\mathbf{E}\left[\left\|\tilde{W}(i)\right\|^2\right]=\mathbf{E}\left[\left\|\tilde{W}(i-1)\right\|^2\right]-2\eta\mathbf{E}\left[e_a(i)f(e(i))\right]+\eta^2\mathbf{E}\left[\left\|X(i)\right\|^2 f^2(e(i))\right] \tag{40}$$

*A. Mean Square Stability*

From (40), if the step-size $\eta$ is chosen such that for all $i$

$$\begin{aligned}\eta &\le \frac{2\mathbf{E}\left[e_a(i)f(e(i))\right]}{\mathbf{E}\left[\left\|X(i)\right\|^2 f^2(e(i))\right]}\\ &=\frac{2\mathbf{E}\left[e_a(i)\exp\left(-\lambda\left|e(i)\right|^\alpha\right)\left|e(i)\right|^{\alpha-1}sign(e(i))\right]}{\mathbf{E}\left[\left\|X(i)\right\|^2\exp\left(-2\lambda\left|e(i)\right|^\alpha\right)\left|e(i)\right|^{2(\alpha-1)}\right]}\end{aligned} \tag{41}$$

then the sequence of weight error power $\mathbf{E}\left[\left\|\tilde{W}(i)\right\|^2\right]$ will be decreasing and converging. According to the analysis results presented in [33], the following theorem holds.

*Theorem 6:* Assume that the noise sequence $\{v(i)\}$ is i.i.d. and independent of $X(i)$, and the filter is long enough such that $e_a(i)$ is Gaussian. Then, for the GMCC algorithm (31), the weight error power $\mathbf{E}\left[\left\|\tilde{W}(i)\right\|^2\right]$ will be monotonically decreasing if the step-size $\eta$ satisfies

$$\eta\le\frac{2}{\left(\mathbf{E}\left[\left\|X(i)\right\|^4\right]\right)^{1/2}}\left(\inf_{\mathbf{E}\left[e_a^2\right]\in\Pi}\frac{\mathbf{E}\left[e_a(i)\exp\left(-\lambda\left|e(i)\right|^\alpha\right)\left|e(i)\right|^{\alpha-1}sign(e(i))\right]}{\sqrt{\mathbf{E}\left[\exp\left(-4\lambda\left|e(i)\right|^\alpha\right)\left|e(i)\right|^{4(\alpha-1)}\right]}}\right) \tag{42}$$

where $\Pi = \left\{ \mathbf{E}\left[ e_a^2 \right] : \xi \le \mathbf{E}\left[ e_a^2 \right] \le \frac{1}{4} Tr\left( \mathbf{R}_{XX} \right) \mathbf{E}\left[ \left\| \tilde{W}(0) \right\|^2 \right] \right\}$, in which $\xi$ is a Cramer-Rao lower bound defined in [33].

To better understand the mean-square stability of the GMCC, it's more important to investigate the probability of divergence (POD) [34]. Here the divergence means $\lim_{i \to \infty} \left\| \tilde{W}(i) \right\|^2 = \infty$ in a realization of an adaptive algorithm. For the LMF algorithm, the POD is nonzero when the input distribution has infinite support, no matter how small the step-size is chosen [34]. Below we present a simple example to show that the GMCC is rather stable and may have zero POD, no matter what input distribution. Let's consider the scalar filtering case [34], in which the desired signal is $d(i) = W_0 X(i)$, where $W_0$ and $X(i)$ are both scalars, and the noise $v(i)$ is assumed to be zero for simplicity. Then, we have

$$\begin{aligned} \tilde{W}(i) &= \tilde{W}(i-1) - \eta f\left(e(i)\right) X(i) \\ &= \left[ 1 - \eta \exp\left( -\lambda \left| e(i) \right|^{\alpha} \right) \left| e(i) \right|^{\alpha - 2} X^2(i) \right] \tilde{W}(i-1) \\ &= \left[ 1 - \eta \exp\left( -\lambda \left| e(i) \right|^{\alpha} \right) \left| e(i) \right|^{\alpha} \left| \tilde{W}(i-1) \right|^{-2} \right] \tilde{W}(i-1) \end{aligned} \tag{43}$$

It is easy to show

$$0 \le \exp\left( -\lambda \left| e(i) \right|^{\alpha} \right) \left| e(i) \right|^{\alpha} \le \frac{1}{\lambda} \exp(-1) \tag{44}$$

So it holds that if $\left| \tilde{W}(i-1) \right|^2 \ge \frac{\eta}{2\lambda} \exp\left(-1\right)$, then

$$0 \le \eta \exp\left( -\lambda \left| e(i) \right|^{\alpha} \right) \left| e(i) \right|^{\alpha} \left| \tilde{W}(i-1) \right|^{-2} \le 2 \tag{45}$$

In this case, we have $\left| 1 - \eta \exp\left( -\lambda \left| e(i) \right|^{\alpha} \right) \left| e(i) \right|^{\alpha} \left| \tilde{W}(i-1) \right|^{-2} \right| \le 1$, and

$$\begin{aligned} \left| \tilde{W}(i) \right|^2 &= \left| 1 - \eta \exp\left( -\lambda \left| e(i) \right|^{\alpha} \right) \left| e(i) \right|^{\alpha} \left| \tilde{W}(i-1) \right|^{-2} \right|^2 \times \left| \tilde{W}(i-1) \right|^2 \\ &\le \left| \tilde{W}(i-1) \right|^2 \end{aligned} \tag{46}$$

Therefore, the limit (if exists) of $\left|\tilde{W}(i)\right|$ always satisfies $\lim_{i\to\infty}\left|\tilde{W}(i)\right| < \infty$, which implies that in this simple example the GMCC will never diverge (or its POD is zero).

When $W(i)$ is a vector and there is a noise $v(i)$, the analysis of the POD of GMCC is very complicated and is left open in this work. However, our simulation results suggest that in most situations the POD of GMCC is zero, even when the noise signal contains large outliers.

*B. Steady-State Mean Square Performance*

With a similar derivation presented in [35], one can analyze the mean square transient behaviors of the algorithm (31). This is a trivial but quite tedious task since we have to evaluate the expectations $\mathbf{E}\left[e_a(i)\exp\left(-\lambda\left|e(i)\right|^{\alpha}\right)\left|e(i)\right|^{\alpha-1} sign(e(i))\right]$ and $\mathbf{E}\left[\left\|X(i)\right\|^2 \exp\left(-2\lambda\left|e(i)\right|^{\alpha}\right)\left|e(i)\right|^{2(\alpha-1)}\right]$. In the following, we only analyze the steady-state mean square performance by using the Taylor expansion method [23].

As the filter reaches the steady-state, we have $\mathbf{E}\left[\left\|\tilde{W}(i)\right\|^2\right] = \mathbf{E}\left[\left\|\tilde{W}(i-1)\right\|^2\right]$. By (40), it holds that

$$2\mathbf{E}\left[e_a(i) f(e(i))\right] = \eta\mathbf{E}\left[\left\|X(i)\right\|^2 f^2(e(i))\right] \tag{47}$$

Assume that $\left\|X(i)\right\|^2$ is asymptotically uncorrelated with $f^2(e(i))$ ( the rationality of this assumption has been discussed in [33]). Then (47) becomes

$$2\mathbf{E}\left[e_a(i) f(e(i))\right] = \eta Tr\left(\mathbf{R}_{XX}\right)\mathbf{E}\left[f^2(e(i))\right] \tag{48}$$

In the steady-state, the distributions of $e_a(i)$ and $e(i)$ are independent of $i$, thus one can omit the time index and simply write (48) as

$$2\mathbf{E}\left[e_a f(e)\right] = \eta Tr\left(\mathbf{R}_{XX}\right)\mathbf{E}\left[f^2(e)\right] \tag{49}$$

Let $S=\lim_{i\to\infty}\mathbf{E}\left[e_a^2(i)\right]=\mathbf{E}\left[e_a^2\right]$ be the steady-state *excess mean square error* (EMSE). An approximate analytical expression of $S$ can be derived . Before proceeding we give two common assumptions:

**A1**: The noise $v(i)$ is zero-mean, independent, identically distributed, and is independent of the input.

**A2**: The *a priori* error $e_a(i)$ is zero-mean and independent of the noise.

Taking the Taylor expansion of $f(e)$ with respect to $e_a$ around $v$, we obtain

$$f(e)=f(e_a+v)=f(v)+f'(v)e_a+\frac{1}{2}f''(v)e_a^2+o(e_a^2) \tag{50}$$

where $o(e_a^2)$ denotes the third and higher-order terms, and

$$f'\left(v\right)=\exp\left(-\lambda|v|^{\alpha}\right)|v|^{\alpha-2}\left((\alpha-1)-\lambda\alpha|v|^{\alpha}\right) \tag{51}$$

$$f''\left(v\right)=\exp\left(-\lambda|v|^{\alpha}\right)sign(v)\left\{-\lambda\alpha\left((\alpha-1)|v|^{2\alpha-3}-\lambda\alpha|v|^{3\alpha-3}\right)+\left((\alpha-1)(\alpha-2)|v|^{\alpha-3}-\lambda\alpha(2\alpha-2)|v|^{2\alpha-3}\right)\right\} \tag{52}$$

If $\mathbf{E}\left[o(e_a^2)\right]$ is small enough, then based on the assumptions A1 and A2 we can derive

$$\mathbf{E}\left[e_a f(e)\right]=\mathbf{E}\left[e_a f(v)+f'(v)e_a^2+o(e_a^2)\right]\approx\mathbf{E}\left[f'(v)\right]S \tag{53}$$

$$\mathbf{E}\left[f^2(e)\right]\approx\mathbf{E}\left[f^2(v)\right]+\mathbf{E}\left[f(v)f''(v)+\left|f'(v)\right|^2\right]S \tag{54}$$

Substituting (53) and (54) into (49), we obtain

$$S=\frac{\eta Tr(\mathbf{R}_{XX})\mathbf{E}\left[f^2(v)\right]}{2\mathbf{E}\left[f'(v)\right]-\eta Tr(\mathbf{R}_{XX})\mathbf{E}\left[f(v)f''(v)+\left|f'(v)\right|^2\right]} \tag{55}$$

Further, substituting (51) and (52) into (55) yields

$$S=\frac{\eta Tr(\mathbf{R}_{XX})\mathbf{E}\left[\exp\left(-2\lambda|v|^{\alpha}\right)|v|^{2\alpha-2}\right]}{2\mathbf{E}\left[\exp\left(-\lambda|v|^{\alpha}\right)|v|^{\alpha-2}\left((\alpha-1)-\lambda\alpha|v|^{\alpha}\right)\right]-\eta Tr(\mathbf{R}_{XX})\mathbf{E}\left[\zeta(v)\right]} \tag{56}$$

where

$$\begin{aligned}\zeta(v) &= \left[ f(v) f''(v) + \left| f'(v) \right|^2 \right] \\ &= \exp\left(-2\lambda |v|^{\alpha}\right) |v|^{2\alpha-4} \left[ (\alpha-1)(2\alpha-3) - 5\lambda\alpha(\alpha-1)|v|^{\alpha} + 2\lambda^2\alpha^2 |v|^{2\alpha} \right]\end{aligned} \tag{57}$$

When the step-size $\eta$ is small enough, (57) can be simplified to

$$S \approx \frac{\eta Tr(\mathbf{R}_{XX})\mathbf{E}\left[\exp\left(-2\lambda |v|^{\alpha}\right)|v|^{2\alpha-2}\right]}{2\mathbf{E}\left[\exp\left(-\lambda |v|^{\alpha}\right)|v|^{\alpha-2}\left((\alpha-1)-\lambda\alpha |v|^{\alpha}\right)\right]} \tag{58}$$

*Remark 7*: Given a noise distribution, one can evaluate the expectations in (56) and obtain a theoretical value of the steady-state EMSE. It is, however, worth noting that the steady-state EMSE of (56) is derived under the assumption that the steady-state *a priori* error $e_a$ is small such that its third and higher-order terms are negligible. When the step-size or noise power is too large, the *a priori* error will also be large. In this case, the derived EMSE value will not accurately enough characterize the performance.

## VI. Simulation Results

Now we present simulation results to confirm the theoretical predictions and demonstrate the desirable performance of the proposed GMCC algorithm.

First, we investigate the stability problem of the GMCC algorithm. Note that the steady-state performance is valid only when the algorithm does not diverge. In many cases, however, an adaptive algorithm may diverge especially at the initial convergence stage. Below we present some simulation results about the probability of divergence (POD) of the GMCC (with $\alpha = 4.0$), compared with that of the LMF algorithm, whose probability of divergence has been studied in [34]. The weight vector of the unknown system is assumed to be $W_0$ =[0.1,0.2,0.3,0.4,0.5,0.4,0.3,0.2,0.1], and the initial weight vector of the adaptive filter is a null vector. The input signal and the disturbance noise are both zero-mean Gaussian with variance 1.0. The PODs with different step-sizes are illustrated in Fig. 2. To evaluate the PODs, 1000 independent Monte Carlo simulations were performed and in each simulation, 1000 iterations were run. We labeled a learning curve as "diverging" if at the last iteration the weight error power $\|W_0 - W(i)\|^2$ is larger than 100. As one can see clearly, compared with the LMF, the GMCC is rather stable and does not diverge at all, and this coincides with our theoretical expectation.

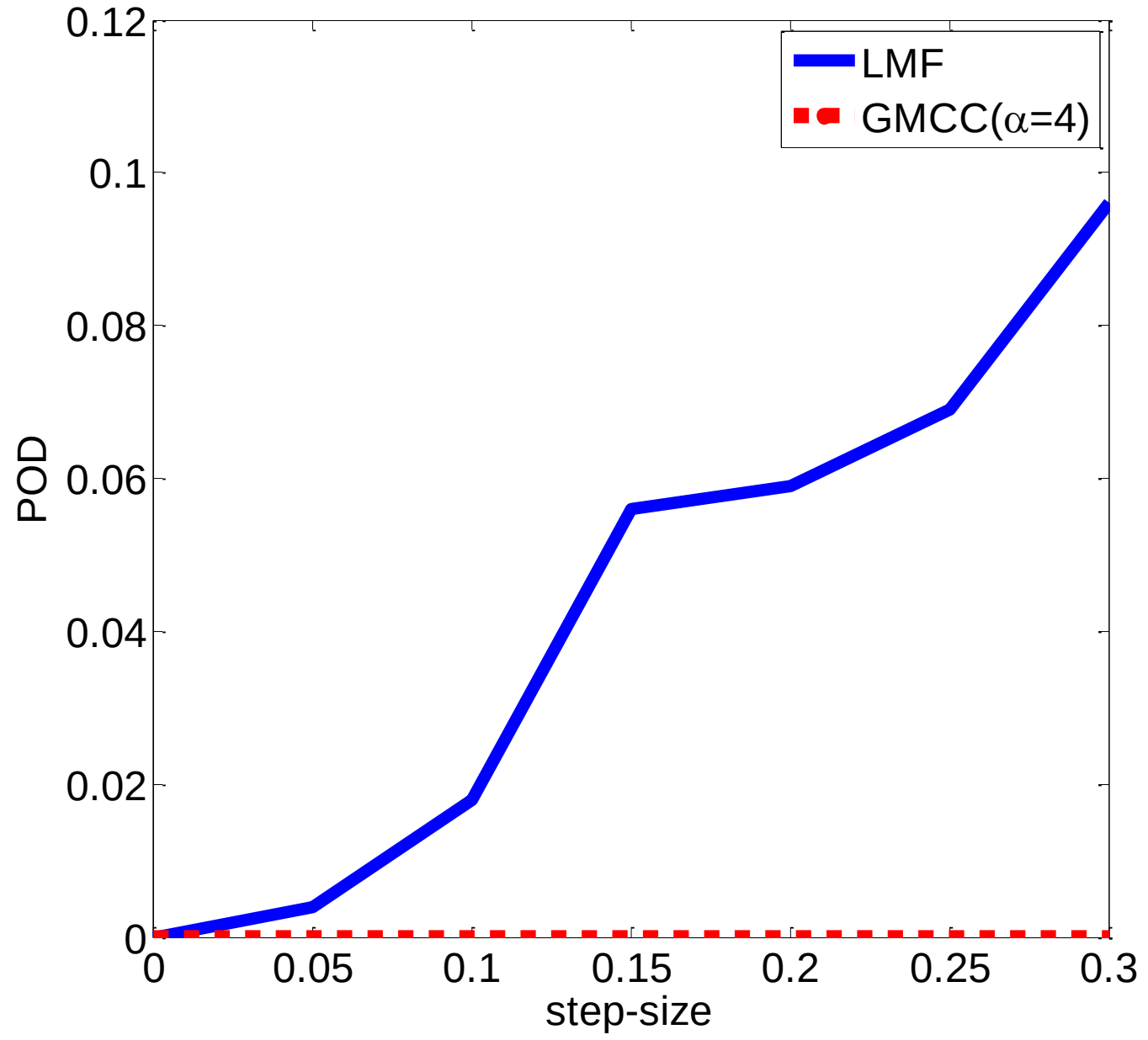

Fig.2 PODs with different step-sizes

Second, we show the theoretical and simulated steady-state performance of the GMCC. In the simulation, we set $\alpha = 4$, and $\lambda = 0.03$. The filter length is 20, the input signal is a zero-mean white Gaussian process with variance 1.0, and the disturbance noise is assumed to be zero-mean and uniform distributed over $\left[-\sqrt{3}, \sqrt{3}\right]$. Fig. 3 shows the steady-state EMSEs with different step-sizes and the noise variances, where the simulated EMSEs are computed as an average over 100 independent Monte Carlo simulations, and in each simulation, 50000 iterations were run to ensure the algorithm to reach the steady state, and the steady-state EMSE was obtained as an average over the last 1000 iterations. One can observe: i) the steady-state EMSEs are increasing with step-size and noise variance; ii) when the step-size and noise variance are small, the steady-state EMSEs computed by simulations match very well the theoretical values computed by (56) ; iii) when the step-size and noise variance become large, the experimental results will, however, gradually differ from the theoretical values, and this also coincides with the theoretical prediction.

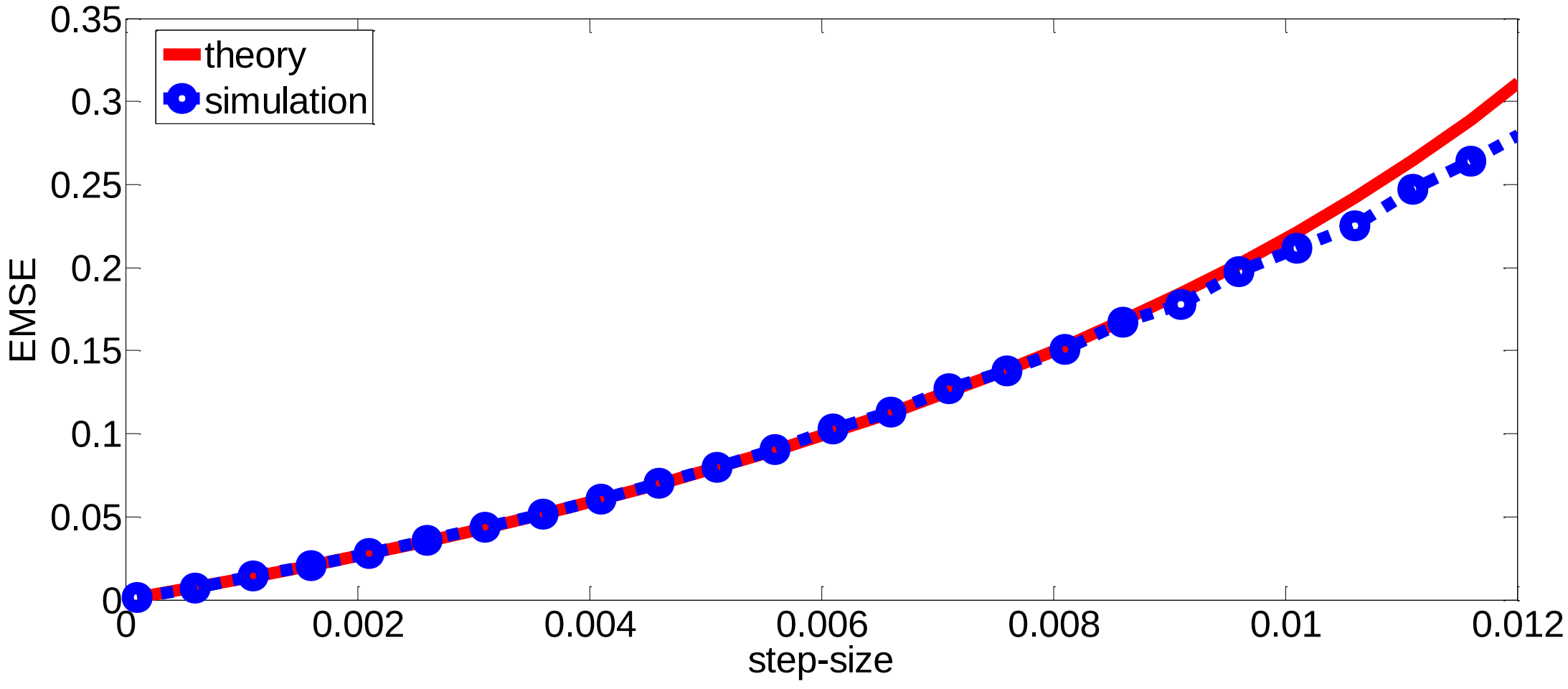

(a)

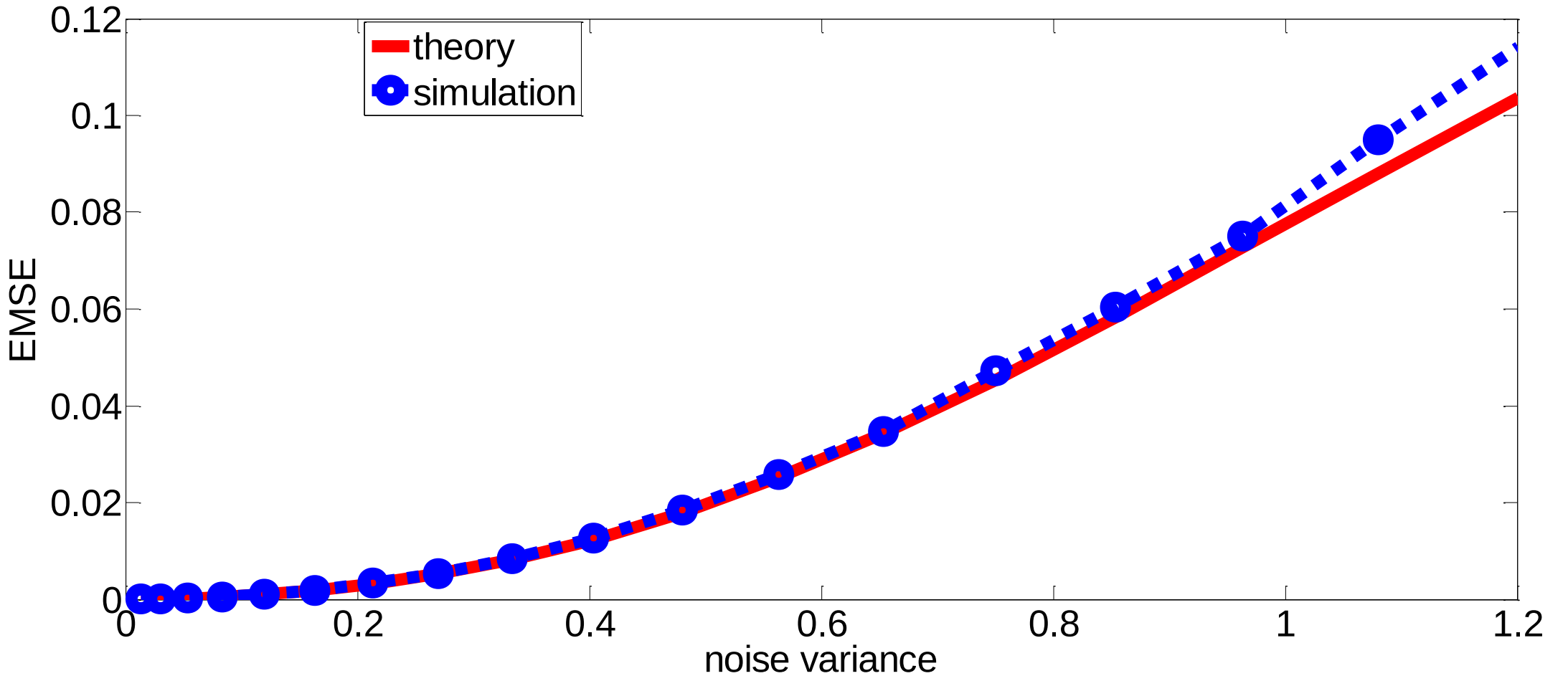

(b)

Fig.3 Theoretical and simulated EMSEs: (a) with different step-sizes; (b) with different noise variances

Third, we compare the performance of the GMCC and the LMP family algorithms with different $p$ values, namely SA ( $p=1$ ) , LMS ( $p=2$ ), and the LMF ( $p=4$ ). The unknown system is assumed to be the same as that in the first simulation. In particular, we consider a noise model with form $v(i)=(1-a(i))A(i)+a(i)B(i)$ , where $a(i)$ is a binary independent and identically distributed process with $\Pr\{a(i)=1\}=c$, $\Pr\{a(i)=0\}=1-c$, and $0 \le c \le 1$ is an occurrence probability; whereas $A(i)$ is a noise process with smaller variance, and $B(i)$ is another noise process with substantially much larger variance to represent large outliers (or impulsive disturbances). The noise processes $A(i)$ and $B(i)$ are mutually independent and they are both independent of $a(i)$. In the simulation, $c$ is set at 0.06, and $B(i)$ is a white Gaussian process with zero-mean and variance 15. For the noise $A(i)$, we consider four distributions: a) Gaussian distribution with zero-mean and unit variance; b) Binary distribution over {-1,1} with probability mass Pr{x=-1}=Pr{x=1}=0.5; c) Laplace distribution with zero-mean and unit variance; d) Uniform distribution over $\left[-\sqrt{3},\sqrt{3}\right]$. The convergence curves in terms of the weight error power $\left\|W_0 - W(i)\right\|^2$ averaged over 100 independent Monte Carlo runs are shown in Fig. 4. In the simulation, the step-sizes are chosen such that all the algorithms have almost the same initial convergence speed. The parameter $\lambda$ in GMCC is experimentally chosen such that the algorithm achieves desirable results. From the simulation results we can observe: 1) the GMCC family algorithms are much more stable (robust) than the LMP family algorithms (In this example, when $p > 4$, the LMP will not converge); 2) the GMCC with $\alpha \neq 2$ may outperform significantly the original MCC ( $\alpha = 2$ ) algorithm. In particular, the GMCC with $\alpha = 6$ achieves the best performance when $A(i)$ is of Binary or Uniform distribution. In order to further demonstrate the robustness of the GMCC against large outliers, we increase the variance of the outlier noise $B(i)$ from 15 to 100. In this case, the LMP family algorithms, except SA and LMS, will diverge, while the GMCC family algorithms can still work well. We show in Fig. 5 the simulation results when $A(i)$ is Uniform distributed.

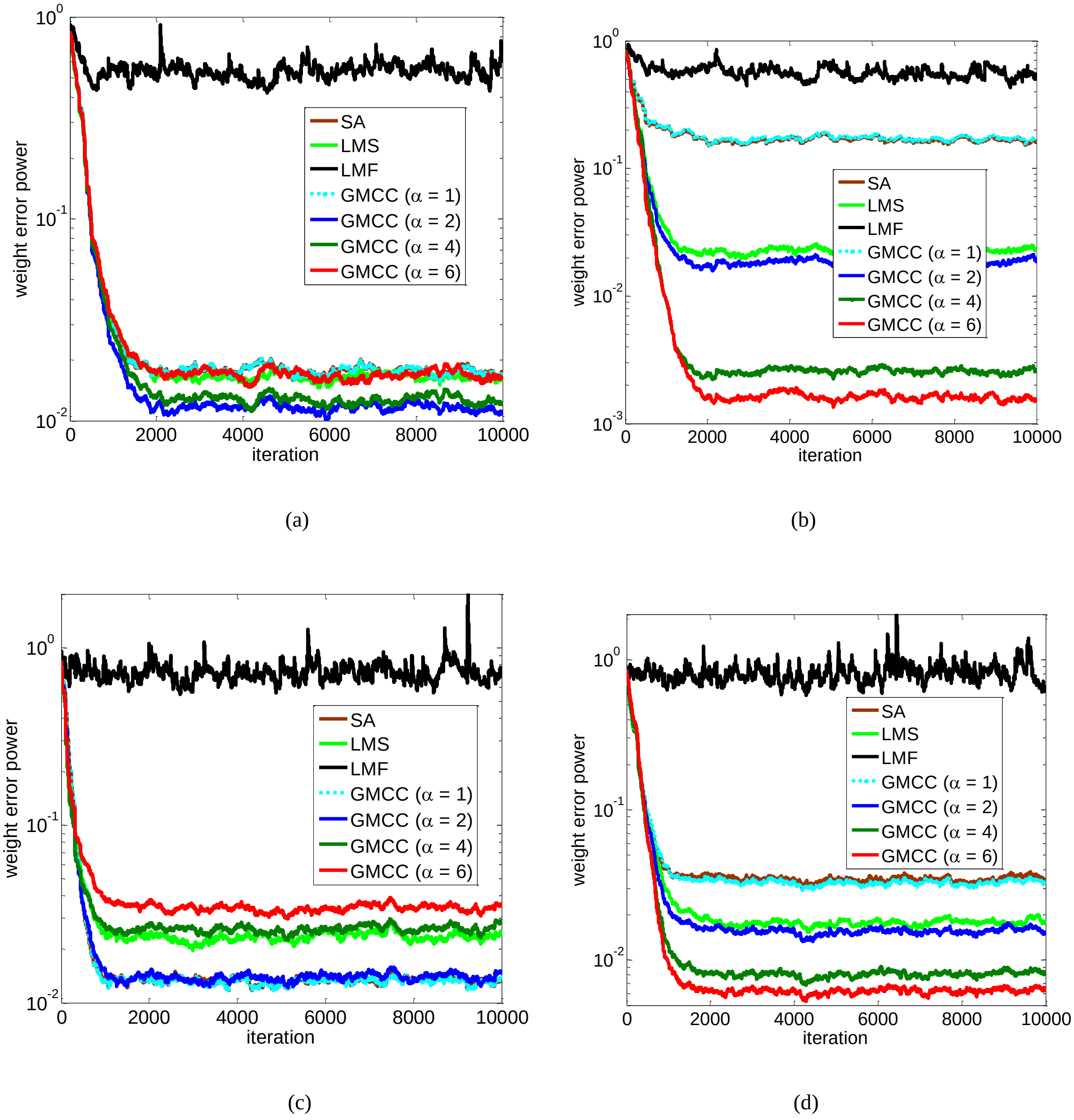

Fig.4 Convergence curves with different distributions of $A(i)$ : (a) Gaussian; (b) Binary; (c) Laplace; (d) Uniform

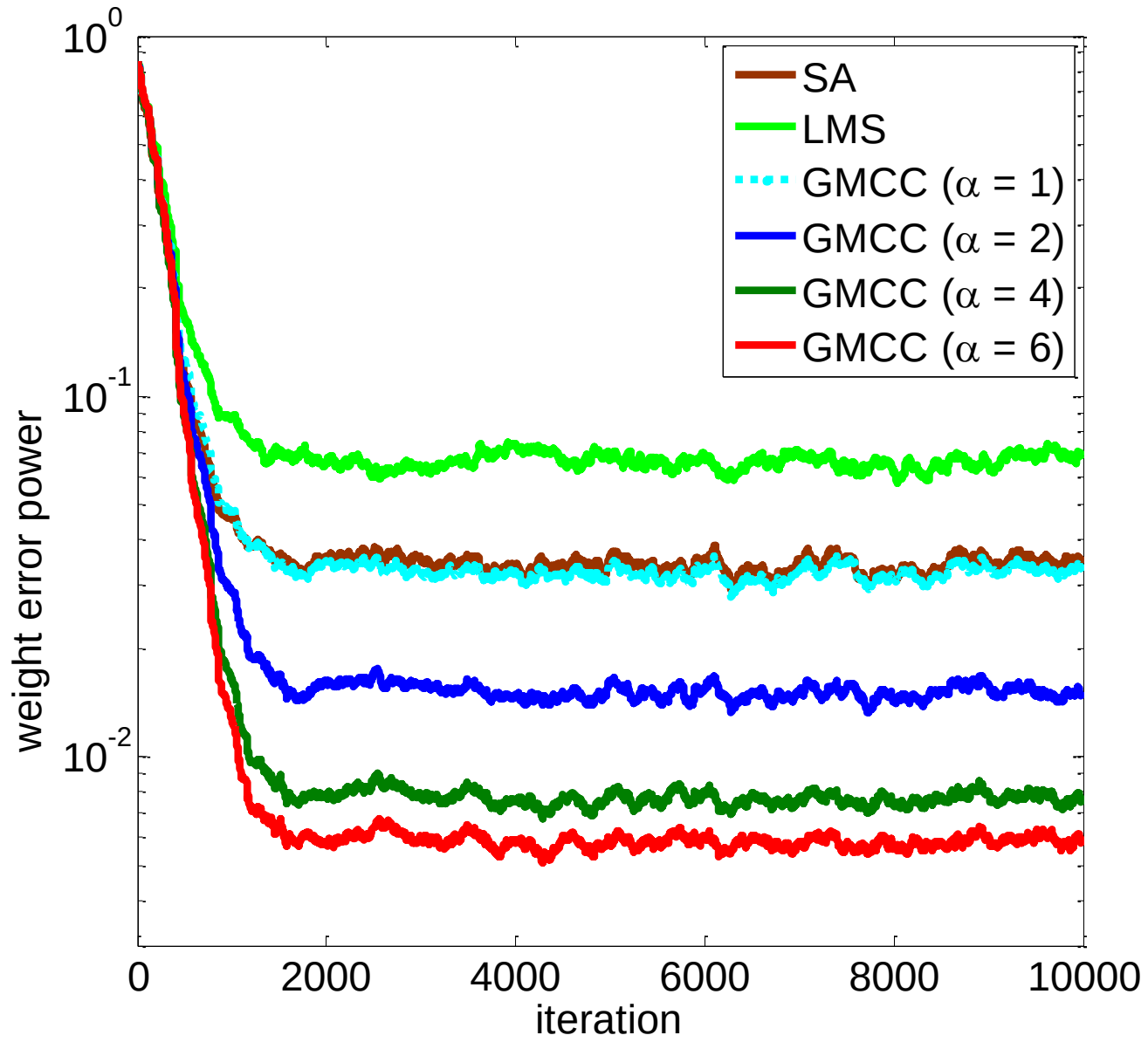

Fig.5 Convergence curves with larger outliers

## VII. CONCLUSION

Correntropy is a novel nonlinear similarity measure in kernel space, and the maximum correntropy criterion (MCC) has recently been widely applied in domains of machine learning and signal processing. In previous studies, the kernel function in correntropy is however limited to the Gaussian kernel without mentioned otherwise. Gaussian kernel is desirable in many cases but obviously, it is not always the best choice. In this work, we proposed a generalized correntropy, using the generalized Gaussian density (GGD) function as the kernel (not necessarily a Mercer kernel). The new definition is very general and flexible, which includes the original correntropy with Gaussian kernel as a special case. Some important properties of the generalized correntropy were presented. The generalized maximum correntropy criterion (GMCC) was also proposed as an optimality criterion in estimation related problems. In particular, we applied the GMCC criterion to adaptive filtering. The optimal solution under GMCC and an adaptive algorithm, called the GMCC algorithm, were derived. Further, we investigated the mean square convergence behaviors of the developed algorithm. A simple example was presented to show that the GMCC algorithm is very stable and will have a zero probability of divergence (POD). A theoretical value of the steady-state *excess mean square error* (EMSE) was also derived. Theoretical results and excellent performance of the GMCC were confirmed by Monte Carlo simulation results.

## APPENDIX A

## PROOF OF PROPERTY 6

*Proof*: When $0 < \alpha \le 2$, we can construct a nonlinear mapping: $\Phi_{\alpha,\beta}\left(\vec{X}\right) = \left[\varphi_{\alpha,\beta}(x_1)^T, \cdots, \varphi_{\alpha,\beta}(x_N)^T\right]^T$.

It follows that

$$\begin{aligned} &\left\|\Phi_{\alpha,\beta}\left(\vec{X}\right) - \Phi_{\alpha,\beta}\left(\vec{Y}\right)\right\| \\ &= \sqrt{\left[\Phi_{\alpha,\beta}\left(\vec{X}\right) - \Phi_{\alpha,\beta}\left(\vec{Y}\right)\right]^T \left[\Phi_{\alpha,\beta}\left(\vec{X}\right) - \Phi_{\alpha,\beta}\left(\vec{Y}\right)\right]} \\ &= \sqrt{2N\left(\mathrm{G}_{\alpha,\beta}(0) - \hat{\mathrm{V}}_{\alpha,\beta}(X,Y)\right)} \\ &= \sqrt{2N} \times GCIM\left(\vec{X},\vec{Y}\right) \end{aligned} \tag{A. 1}$$

which implies that the GCIM function $GCIM\left(\vec{X},\vec{Y}\right)$ defines a "*Euclidean distance*" in the Hilbert space $\mathscr{F}_\kappa^N$. Hence, $GCIM\left(\vec{X},\vec{Y}\right)$ is a metric in the sample vector space since it satisfies: i) $GCIM\left(\vec{X},\vec{Y}\right) \ge 0$; ii) $GCIM\left(\vec{X},\vec{Y}\right) = GCIM\left(\vec{Y},\vec{X}\right)$; iii) $GCIM\left(\vec{X},\vec{Z}\right) \le GCIM\left(\vec{X},\vec{Y}\right) + GCIM\left(\vec{Y},\vec{Z}\right)$.

## APPENDIX B

## PROOF OF PROPERTY 7

*Proof*: As $\lambda \to 0+$ (or $x_i \to 0, i = 1, \cdots, N$), we have

$$\begin{aligned} L_{\alpha,\beta}\left(\vec{X}\right) &= \left(\frac{N}{\lambda\gamma_{\alpha,\beta}}\hat{J}_{GC-loss}(X,0)\right)^{1/\alpha} \\ &= \left[\frac{N}{\lambda\gamma_{\alpha,\beta}}\left(\mathrm{G}_{\alpha,\beta}(0)-\frac{1}{N}\sum_{i=1}^{N}\mathrm{G}_{\alpha,\beta}(x_i)\right)\right]^{1/\alpha} \\ &\approx \left[\frac{N}{\lambda\gamma_{\alpha,\beta}}\left(\gamma_{\alpha,\beta}-\frac{1}{N}\sum_{i=1}^{N}\gamma_{\alpha,\beta}\left(1-\lambda\left|x_i\right|^{\alpha}\right)\right)\right]^{1/\alpha} \\ &= \left[\sum_{i=1}^{N}\left|x_i\right|^{\alpha}\right]^{1/\alpha} \end{aligned} \tag{B. 1}$$

## APPENDIX C

## PROOF OF PROPERTY 8

*Proof*: Let $\vec{X}_0$ be the solution obtained by minimizing $\left\|\vec{X}\right\|_0$ over $\Omega$ and $\vec{X}_l$ the solution achieved by minimizing $L_{\alpha,\beta}\left(\vec{X}\right)$. Then $L_{\alpha,\beta}\left(\vec{X}_l\right)\le L_{\alpha,\beta}\left(\vec{X}_0\right)$, and hence

$$\sum_{i=1}^{N}\mathrm{G}_{\alpha,\beta}\left((\vec{X}_l)_i\right)\ge\sum_{i=1}^{N}\mathrm{G}_{\alpha,\beta}\left((\vec{X}_0)_i\right) \tag{C. 1}$$

where $(\vec{X}_l)_i$ denotes the $i$ th component of $\vec{X}_l$. It follows that

$$\begin{aligned} &\left(N-\left\|\vec{X}_l\right\|_0\right)+\sum_{i=1,(\vec{X}_l)_i\neq 0}^{N}\exp\left(-\lambda\left|(\vec{X}_l)_i\right|^{\alpha}\right) \\ &\qquad\ge\left(N-\left\|\vec{X}_0\right\|_0\right)+\sum_{i=1,(\vec{X}_0)_i\neq 0}^{N}\exp\left(-\lambda\left|(\vec{X}_0)_i\right|^{\alpha}\right) \end{aligned} \tag{C. 2}$$

Thus we have

$$\left\|\vec{X}_l\right\|_0-\left\|\vec{X}_0\right\|_0\le\sum_{i=1,(\vec{X}_l)_i\neq 0}^{N}\exp\left(-\lambda\left|(\vec{X}_l)_i\right|^{\alpha}\right)-\sum_{i=1,(\vec{X}_0)_i\neq 0}^{N}\exp\left(-\lambda\left|(\vec{X}_0)_i\right|^{\alpha}\right) \tag{C. 3}$$

Since $\left|x_i\right|>\delta$, $\forall i: x_i\neq 0$, as $\lambda\to\infty$ the right hand side of (C. 3) will approach zero. Therefore, if $\lambda$ is large enough, it holds that

$$\left\|\vec{X}_0\right\|_0 \le \left\|\vec{X}_l\right\|_0 \le \left\|\vec{X}_0\right\|_0 + \varepsilon \tag{C. 4}$$

where $\varepsilon$ is a small positive number arbitrarily close to zero.

## APPENDIX D

## PROOF OF PROPERTY 9

*Proof*: The Hessian (if exists) of $\hat{J}_{GC-loss}$ with respect to $\vec{e}$ is

$$\mathbf{H}_{\hat{J}_{GC-loss}}\left(\vec{e}\right) = -\frac{\alpha\lambda\gamma_{\alpha,\beta}}{N}\begin{pmatrix} T(e_1)\left(\alpha\lambda\left|e_1\right|^{\alpha} - (\alpha-1)\right) & 0 & \cdots & 0 \\ 0 & T(e_2)\left(\alpha\lambda\left|e_2\right|^{\alpha} - (\alpha-1)\right) & \cdots & 0 \\ \vdots & \vdots & \ddots & \vdots \\ 0 & 0 & \cdots & T(e_N)\left(\alpha\lambda\left|e_N\right|^{\alpha} - (\alpha-1)\right) \end{pmatrix} \tag{D. 1}$$

where $T(x) = \exp\left(-\lambda\left|x\right|^{\alpha}\right)\left|x\right|^{\alpha-2}$. From (D. 1) one can see:

i) if $0 < \alpha \le 1$, then $\mathbf{H}_{\hat{J}_{GC-loss}}\left(\vec{e}\right) \le \mathbf{0}$ for any $\vec{e}$ with $e_i \ne 0$ ( $i = 1,\cdots,N$ ) ;

ii) if $\alpha > 1$, then $\mathbf{H}_{\hat{J}_{GC-loss}}\left(\vec{e}\right) \ge \mathbf{0}$ for any $\vec{e}$ with $0 < \left|e_i\right| \le \left[(\alpha-1)/\alpha\lambda\right]^{1/\alpha}$ ( $i = 1,\cdots,N$ );

iii) if $\lambda \to 0+$, then for any $\vec{e}$ with $e_i \ne 0$ ( $i = 1,\cdots,N$ ), we have $\mathbf{H}_{\hat{J}_{GC-loss}}\left(\vec{e}\right) \le \mathbf{0}$ for $0 < \alpha \le 1$, and $\mathbf{H}_{\hat{J}_{GC-loss}}\left(\vec{e}\right) \ge \mathbf{0}$ for $\alpha > 1$.

## APPENDIX E

## PROOF OF PROPERTY 10

*Proof*: A differentiable function $f : S \mapsto \mathbb{R}$ ( $S \subset \mathbb{R}^N$ ) is said to be invex, if and only if []

$$f(x_2) \ge f(x_1) + q(x_1, x_2)^T \nabla f(x_1), \ \forall x_1, x_2 \in S \tag{E. 1}$$

where $\nabla f(x)$ denotes the gradient of $f$ with respect to $x$, and $q(x_1, x_2)$ is some vector valued function. For $\alpha > 1$, the GC-loss $\hat{J}_{GC-loss}$ is a differentiable function of $\vec{e}$, and the gradient $\nabla\hat{J}_{GC-loss}(\vec{e})$ is

$$\nabla\hat{J}_{GC-loss}(\vec{e}) = \frac{\lambda\alpha\gamma_{\alpha,\beta}}{N}\left[\exp\left(-\lambda|e_1|^{\alpha}\right)|e_1|^{\alpha-1} sign(e_1) \quad \cdots \quad \exp\left(-\lambda|e_N|^{\alpha}\right)|e_N|^{\alpha-1} sign(e_N)\right]^T \tag{E. 2}$$

where $sign(.)$ is the sign function. Since $e_i \le M$, we have $\nabla\hat{J}_{GC-loss}(\vec{e}) = \mathbf{0}$ if and only if $\vec{e} = \mathbf{0}$. On the other hand, we have $\hat{J}_{GC-loss}(\vec{e}) \ge \hat{J}_{GC-loss}(\mathbf{0}) = 0$. So we can construct the following function

$$q(\vec{e}_1, \vec{e}_2) = \begin{cases} \dfrac{\hat{J}_{GC-loss}(\vec{e}_2) - \hat{J}_{GC-loss}(\vec{e}_1)}{\nabla\hat{J}_{GC-loss}(\vec{e}_1)^T \nabla\hat{J}_{GC-loss}(\vec{e}_1)} \nabla\hat{J}_{GC-loss}(\vec{e}_1) & if \ \vec{e} \ne \mathbf{0} \\ \mathbf{0} & if \ \vec{e} = \mathbf{0} \end{cases} \tag{E. 3}$$

such that it holds

$$\hat{J}_{GC-loss}(\vec{e}_2) \ge \hat{J}_{GC-loss}(\vec{e}_1) + q(\vec{e}_1, \vec{e}_2)^T \nabla\hat{J}_{GC-loss}(\vec{e}_1) \tag{E. 4}$$

## APPENDIX F

## PROOF OF THEOREM 1

It is easy to derive

$$\begin{aligned} \mathrm{V}_{\alpha,\beta}(X, \hat{X}) &= \int_{\mathbb{R}} \mathrm{G}_{\alpha,\beta}(e) \int_{\mathbb{R}^m} p_{X|Y}(e + g(y) \mid y) dF_Y(y) de \\ &= \int_{\mathbb{R}^m} \left\{ \int_{-\infty}^{\infty} \mathrm{G}_{\alpha,\beta}(e) p_{X|Y}(e + g(y) \mid y) de \right\} dF_Y(y) \\ &= \int_{\mathbb{R}^m} \left\{ \int_{-\infty}^{\infty} \mathrm{G}_{\alpha,\beta}(e' - g(y)) p_{X|Y}(e' \mid y) de' \right\} dF_Y(y) \\ &\overset{(a)}{=} \int_{\mathbb{R}^m} \left\{ \int_{-\infty}^{\infty} \mathrm{G}_{\alpha,\beta}(g(y) - e') p_{X|Y}(e' \mid y) de' \right\} dF_Y(y) \\ &= \int_{\mathbb{R}^m} \left\{ \left( \mathrm{G}_{\alpha,\beta}(.) * p_{X|Y}(. \mid y) \right) (g(y)) \right\} dF_Y(y) \\ &= \int_{\mathbb{R}^m} \rho_{\alpha,\beta}(g(y) \mid y) dF_Y(y) \end{aligned} \tag{F. 1}$$

where $e' = e + g(y)$, and (a) comes from the symmetry of $G_{\alpha,\beta}(.)$. Thus

$$
\begin{aligned}
& g_{GMCC} = \arg\max_{g \in \mathbf{G}} \int_{\mathbb{R}^m} \rho_{\alpha,\beta}(g(y) \mid y) dF_Y(y) \\
& \Rightarrow \; g_{GMCC}(y) = \arg\max_{x \in \mathbb{R}} \rho_{\alpha,\beta}(x \mid y), \forall y \in \mathbb{R}^m
\end{aligned} \tag{F. 2}
$$

which completes the proof.

## ACKNOWLEDGEMENTS

This work was supported by 973 Program (No. 2015CB351703) and National NSF of China (No. 61372152).